\documentclass[letterpaper, 10 pt, conference]{ieeeconf}  

\IEEEoverridecommandlockouts                              

\usepackage{graphicx}
\usepackage[acronym]{glossaries}

\makeglossaries

\title{\LARGE \bf
Accelerating Integrated Task and Motion Planning with\\ Neural Feasibility Checking
}

\author{Lei Xu, Tianyu Ren, Georgia Chalvatzaki, Jan Peters
\thanks{Faculty of Computer Science, Technische Universität Darmstadt
}
}

\begin{document}

\maketitle
\thispagestyle{empty}
\pagestyle{empty}
\newacronym{tamp}{TAMP}{Task and Motion Planning}
\newacronym{nfc}{NFC}{neural feasibility classifier}
\newacronym{cnn}{CNN}{convolutional neural network}
\newacronym{idtmp}{IDTMP}{iteratively deepened task and motion planning}

\newcommand{\content}{
   \section{INTRODUCTION}
\label{sec:intro}

The application of autonomous robotics is rapidly increasing in the industrial and manufacturing sectors. However, a long-wished vision is having robots operate in real-world unstructured environments, interact with humans, and execute complex tasks that concern services in houses, nursing homes, hospitals, etc. While robots manifested repetitive tasks in static and structured settings, we are far from achieving the goal of general-purpose autonomous personal robots. One of the many challenges that hinder robots from executing complex tasks is their long planning time. Depending on the combinatorics of the problem that needs to be solved, planning may terminate without a solution \cite{garrett_integrated_2021}. 

While \gls{tamp} offers a framework for optimizing over task action-sequences and feasible geometric motions, it suffers from the logical gap between the symbolic and geometric space \cite{kaelbling_integrated_2013}. Indeed, in the early stages of the \gls{tamp} research, this hierarchical problem was decomposed into two separate sub-problems: first finding an action sequence and then finding continuous parameter values for each action. This strictly hierarchical approach is incomplete, as there are no guarantees that actions selected by the task planner will be geometrically feasible. For example, the Shakey robot executed STRIPS \cite{fikes_strips_1971} to plan a high-level abstract action plan, such as which room to move to, and then planned for the corresponding low-level motion. Unfortunately, if the low-level motion planning was not feasible, there was no mechanism to find an alternative high-level plan. This isolation of the planning problems in practice rarely holds. For example, in Fig. \ref{fig:unpack scene}, the robot has to move the green box to the other region. The straightforward action sequence is to pick up the green box and put it down to the other region. However, the pick-up action is blocked by the other two taller boxes thus, the low-level motion fails.


\begin{figure}[t]
	\centering
	\includegraphics[width=0.9\linewidth]{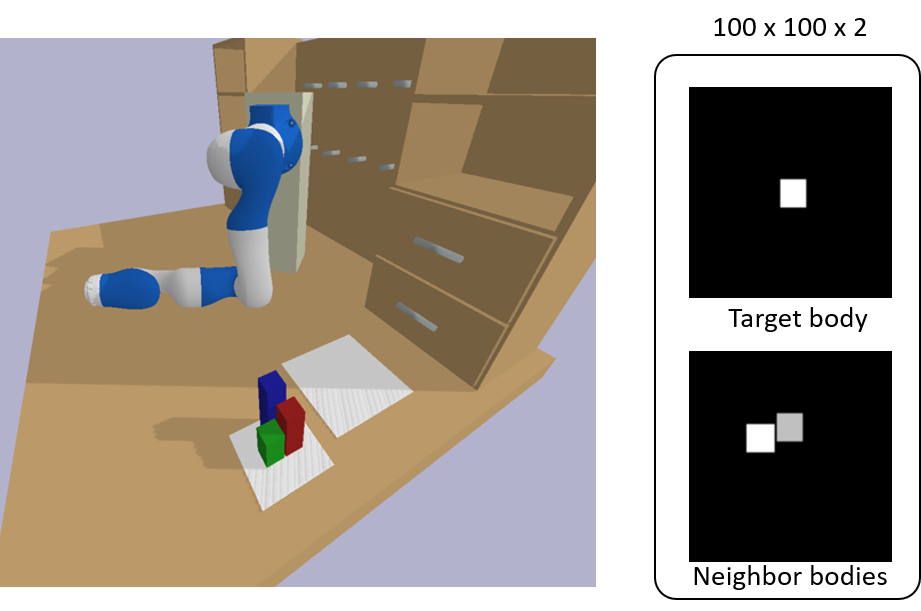}
	\caption{Scene of \textit{Unpack} problem. Left: Three boxes of different sizes are placed together. The goal is to place the green one in the other region. The robot is only allowed to pick a box from the above. TAMP solver must figure out that it has to relocate the two taller boxes away to reach the target green box. Right: We use the local depth image around the target body to predict the feasibility of an action. The depth image is composed of two channels, with the target box in the first channel and other boxes in the second channel.}
	\label{fig:unpack scene}
	\end{figure}

Over the last two decades, researchers naturally opted for extending \gls{tamp} considering the integration of discrete-continuous actions \cite{kaelbling_hierarchical_2011}, leading to a hybrid hierarchical search problem that involves, on the high-level task planning, i.e., the selection of a finite sequence or skeletons of discrete symbolic actions (e.g., which objects to pick and place), on the low-level motion planning, i.e., the selection of parameters in the continuous space to instantiate the sequence (e.g., the pose and grasp of the movable objects, path between two configurations), and in-between an interface for sharing information between discrete and continuous layer, as to trigger the generation of alternate skeletons when the motion generation of the current task plan fails \cite{garrett_integrated_2021,srivastava_combined_2014}. 

Interleaving task planning and motion planning, with the scope of finding a feasible solution, usually involves a large search space that is computationally exhaustive. In particular, the low-level motion planner has to validate the feasibility of the generated action sequences and to instantiate the final kinematically feasible action sequence, while the most action sequences are infeasible, which consumes much of the planning time, making most TAMP methods inefficient and impractical for real world scenario. To speed up the planning process in \gls{tamp} and increase its probability of finding an overall feasible solution in a given time-budget, several heuristics have been introduced to reduce the search space. In particular, in the era of robot learning, different learned heuristics have been integrated in \gls{tamp} to accelerate its decision-making process, and support its probabilistic completeness to find a feasible plan in a specific planning horizon \cite{wells_learning_2019, driess_deep_2020,silver2021learning}.  \cite{driess_deep_2020} introduces a more general visual-based neural network to do the classification of tabletop manipulation. In this paper, we propose a novel CNN-based neural feasibility classifier (NFC) using the local visual information around the target body, object-centered image. With this idea, we are able to extend the feasibility prediction to the entire workspace of robot manipulators. Our model is also able to deal with the case that there is obstacle over the target body.

We implement our work based on the sequencing-first and discretizing-based TAMP methods, iteratively deepened task and motion planning (IDTMP). The main contributions in this work are two-fold,
\begin{itemize}
\item We propose a more general neural feasibility classifier (NFC) that predicts the feasibility of action based on object-centered image of the scene.
\item We show the generalization and high prediction accuracy of the NFC in the entire workspace of the manipulator.
\end{itemize}
   \section{RELTAED WORK}
\label{sec:related_work}

\subsection{TAMP}
\label{sec:rw_tamp}

Theoretically, general-purpose TAMP uses symbolic AI planning \cite{ren_extended_2021} to search in task spaces and can deal with problems with large action space, non-geometric action, etc. Top-down hierarchical planning means also sequencing-first planning \cite{garrett_integrated_2021}. It generates high-level task plan first and maps each action of the plan to the motion space and finds the corresponding motion plan. Sequencing-first planning is more in line with human behavior. In this paper, we only talk about the related works in this direction. Sequencing-first TAMP methods are time efficient while the time-consuming motion planning process is postponed as late as possible \cite{garrett_integrated_2021}.

Due to the inherent difficulty of task planning and motion planning, the former is NP-hard, and the latter is PSPACE hard. There is a compromiss between algorithm performance and algorithm completeness. TAMP methods in the early stage consider only restricted domains to improve the efficient \cite{strip_shakey_1966}, \cite{stilman_navigation_nodate}, \cite{cambon_robot_nodate}. \cite{kaelbling_hierarchical_2011} can guarantee completeness under the assumption of reversible action but at the cost of large amount of time and energy consuming. Therefore, the second challenge of algorithm design for integrated TAMP besides the completeness is to combine task planner and motion planner efficiently. There are two typical strategies to deal with the hybrid planning of the discrete task space and continuous motion space, discretizing-based and sampling-based. \cite{erdem_combining_2011}, \cite{lagriffoul_efficiently_2014}, \cite{lozano-perez_constraint-based_2014}, \cite{dantam_incremental_2018} require the pre-discretization of the continuous space so that the solution space can be interactively pruned according to the motion constraints from motion planning. \cite{srivastava_combined_2014}, \cite{ren_extended_2021}, \cite{garrett_pddlstream_2020} directly sample in the continuous space during the planning process, to guarantee that no potential poses will be missed by the discretization process and show better robustness. However, both strategies need to continuously call low-level motion planner to exclude most infeasible high-level task plans. Motion planner becomes the bottleneck of the original TAMP methods, since it have to be called so many times, to validate the feasibility of candidate task plans.

\subsection{Heuristic for Motion Planning}
\label{sec:rw_heuristics_mp}
In order to improve the performance of motion planning, there are many tricks and techniques being proposed. E-graph \cite{phillips_e-graphs_2012} which is learned from the online planning experience, represents the connectivity of the space and accelerates its planning effort, but it's not clear how it should be adapted to scenes with unknown objects. Lightning \cite{berenson_robot_2012} and Thunder \cite{coleman_experience-based_2014} are two experience-based frameworks using the strategy of experience storing, retrieving, and repairing to speed up high-dimensional motion planning problems. \cite{arslan_machine_2015} proposes a learning-based heuristic to predict the collision-free sample and to estimate the relevant region of the problem. \cite{ichter_learning_2019} proposes a methodology for non-uniform sampling using the sampling distribution learned from demonstration, i.e. successful motion planning, human demo. Our approach differs by using learning to guide the TAMP search and doesn't require any modification of the motion planner. Therefore, above mentioned heuristics for motion planner can also be viewed as a complementary to our method.

\subsection{Learning to Guide TAMP}
\label{sec:rw_heuristics_tamp}
There are also learning-based techniques, which aim to guide the TAMP process. \cite{chitnis_guided_2016} trains heuristics to estimate the difficulty of high-level plan refinement and uses reinforcement learning to propose continuous value for plan refinement in low-level, but human demonstration is required for training and detail about prediction error is not mentioned. “Score-space” is introduced in \cite{kim_learning_2019} as a metric to measure similarity between problem instances to improve motion planning time and transfers knowledge to other problems. Dex-net \cite{mahler_dex-net_2017} can predict the success rate for picking up an object with an arbitrary shape. But the interference of neighbor bodies is not considered. Closely related to our work in this paper, Wells et. al. \cite{wells_learning_2019} trained a feature-based SVM model to help with checking the feasibility of actions instead of using the expensive motion planner. For the same purpose, \cite{driess_deep_2020} validate that it's possible to predict the motion feasibility of a mixed-integer program from visual input with a high accuracy. The learned DVH model can well generalize to scenes with multiple objects and object with different shapes rather than Wells' feature-based SVM model. However, Both ideas are only applicable for a manipulator to operate on a desktop scenario. Our approach will overcome this problem and predict the feasibility of actions in the entire 3D-Cartesian workspace of a specific robot.

   \section{PRELIMINARIES}
\label{sec:preliminary}

\subsection{Task Planning}
\label{sec:pre_tp}
Task planning finds a discrete sequence of actions to transition from the given start state to the desired goal condition. For the general-purpose TAMP problem, PDDL (Planning Domain Definition Language) is the most widely used modeling language to describe the planning domain and a instantiated planning problem. PDDL-domain defines the “universal” aspects of problems. Essentially, these are the aspects that do not change regardless of what specific situation we’re trying to solve. A typical PDDL-domain file contains,
\begin{itemize}
    \item \textbf{predicates} $S$: the symbolic states. The scene state can be described using the combination of predicates.
    \item \textbf{operators} $A$: the symbolic actions, describing the transition of two states, which mainly includes action preconditions and effects after taking the action.
\end{itemize}

We also need a PDDL-problem file to model a specific planning problem. The PDDL-problem describes the following elements:
\begin{itemize}
    \item \textbf{initial states} $s_{0}$ described with the combination of predicates.
    \item \textbf{a set of goal states} $S_g$ described with the combination of predicates.
\end{itemize}

With the PDDL-domain and PDDL-problem defined, there are many task planner $\mathcal{TP}$ in AI planning community available for generating task plans. A task plan $T$ or action sequence $<a_0, ..., a_{n-1}>$ transits the initial state $s_0$ sequentially to one state in the goal state set $S_g$. So the task planning problem can be formulated as (1).
$$
T = <a_0, ..., a_{n-1}> = \mathcal{TP}(S,A,s_0,S_g) \eqno{(1)}
$$
such that for, $ \forall i=0, 1, ... n-1 ,\\ $
$$
    s_i + a_i = s_{i+1} ,
    a_i \in A ,
    s_i \in S ,
    s_n \in S_g 
$$

\subsection{Motion Planning}
\label{sec:pre_mp}
Motion planning for manipulator $\mathcal{MP}$ is to find a motion plan $M$ or configuration sequence $<q_0, ..., q_n>$ which transfers the robot from its initial configuration $q_0$ to the one of the goal configuration $q_n$ without self-collision and collision with the environment, formulated in (2). No collision also means that every configuration of the motion planning solution should be located in the free configuration space of the manipulator $C_{free}$.

$$
    M = <q_0, ..., q_n> = \mathcal{MP}(q_0, q_n, C_{free}) \eqno{(2)}
$$

\subsection{Task and Motion Planning}
\label{sec:pre_tamp}
The key requirement for task planner in TAMP is to generate alternate high-level plans. As TAMP iterates between the task planning and motion planning layers, feedback from the motion planner ideally influences the task planner to generate new plans or to disapprove of geometrically difficult or even infeasible plans. Therefore, the task planning layer must be able to calculate alternative plans and ideally reuse previous planning experience to improve performance.

TAMP combines the discrete action selection of task planning with the continuous path generation of motion planning. The fundamental requirement of the task–motion layer is to build a bridge between task operators and motion planning problems. Given a symbolic task plan, what are the corresponding geometric plan? How should task planner react to the feedback from motion planner? The task–motion interface must translate between the low-level scene geometry and the high-level task descriptions.

We can formulate general TAMP problems as (3)
$$
T, M = \mathcal{TAMP}(S, A, s_0, S_g, C_{free}, q_{0}) \eqno{(3)}
$$

TAMP generates a sequence of actions $T = (a_0, a_1, ..., a_{n-1})$ following a sequence of task states $s_0, s_1, ..., s_n$ and a sequence to motion plans $M_0, M_1, ..., M_{n-1}$, such that for $\forall i= 0, 1, ..., n-1$, \\
$$
    s_n \in S_g,\\
    s_i + a_i = s_{i+1},\\
    a_i \Leftrightarrow M_i,\\
$$
$$
    M_i(0) \in s_i,\\
    M_i(1) \in s_{i+1},\\
    M_i(1) = M_{i+1}(0)\\
$$

\begin{figure}[t]
	\centering
	\includegraphics[width=1\linewidth]{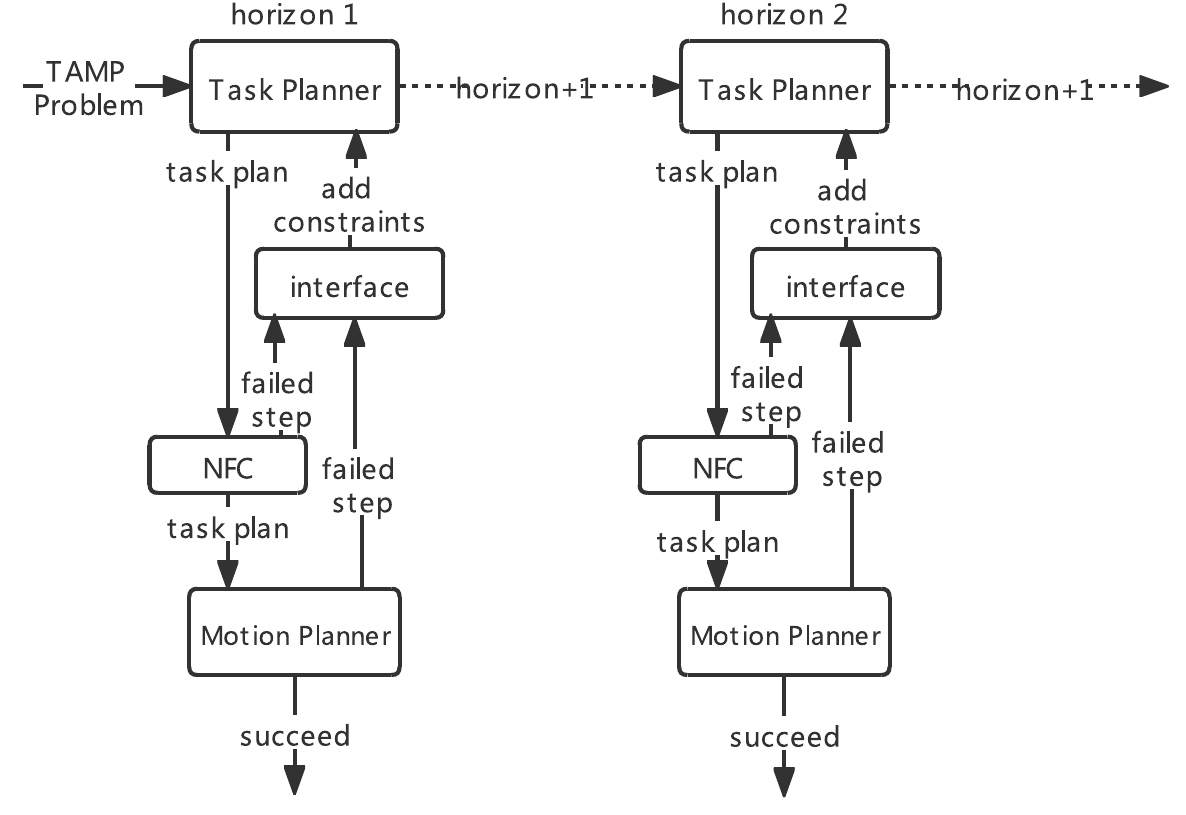}
	\caption{The Workflow of IDTMP integrated with NFC. The original IDTMP doesn't have NFC and motion planner must classify the feasibility of each action sequence by itself. As for IDTMP-NFC, only feasible task plan validated by NFC will be passed to the time-consuming motion planner. As for infeasible task plans, NFC will return its failed step to the interface for updating motion constraints.}
	\label{fig:IDTMP-NFC structure}
\end{figure}

\subsection{Iteratively Deepened Task and Motion Planning (IDTMP)}
\label{sec:pre_idtmp}
Theoretically, our NFC is applicable to all versions of sequencing-first TAMP methods. We take the discretizing-based IDTMP \cite{dantam_incremental_2018} \cite{dantam_task-motion_2018} as the example and integrate our NFC with it. The algorithm workflow of IDTMP is shown in Fig. \ref{fig:IDTMP-NFC structure}. As for its task planner, it employs constraint-based SAT-planner to generate single candidate task plan according to the logical and motion constraints iteratively. Motion planner, such as sampling-based rapidly-exploring random tree(RRT) will validate the feasibility of each action in the candidate task plan. When it's infeasible, motion planner will return failed step so that the interface layer can add extra motion constraints to SAT-planner, so to prune the symbolic solution space. When feasible, motion planner will find the corresponding geometric motion for each action, such as inverse kinematics and path in configuration space. IDTMP searches and validates plans incrementally or horizon by horizon. When task and motion planning on the short horizon failed, IDTMP will jump onto longer horizons and try more complex task plans. To ensure the algorithm completeness, IDTMP will clear all collected motion constraints on the last horizon, so that previously failed task plans could be retrospected. TAMP process come to end until both task and motion plan are found or the allotted planning time is exceeded.


   \section{NEURAL FEASIBILITY CLASSIFIER}
\label{sec:nfc}
The goal of our work is to train a classifier which can predict the action feasibility in the 3d-Cartesian workspace of a specific robot arm. In order to realize this goal, the key requirement is how to encode involved elements, including the target object of different shape, the environment and action type. Driess et. al. \cite{driess_deep_2020}, \cite{driess_reasoning_2020} have shown the effectiveness of visual heuristics on desktop scenarios. Our heuristic is also visual-based but applies a more efficient way to encode the scene.

\begin{figure}[t]
    \centering
    \includegraphics[width=1\linewidth]{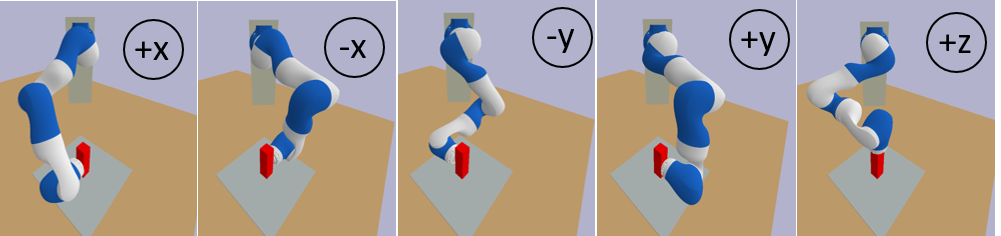}
    \caption{Visualization of 5 different discrete actions. Different actions means it has different pick-up or put-down directions, which are +x, -x, +y, -y, +z direction of the body frame.}
    \label{fig:visualiation pick direction}
\end{figure}

Our neural feasibility classifier employs a local depth image (Fig. \ref{fig:unpack scene}) which captures the neighboring area (boundary of 0.5m x 0.5m) around the target body, object-centered image. The neighboring area of target body means, the area around to be picked object for \textit{pick-up} action, and the area around the placing pose of grasped body for \textit{put-down} action. The image pixel out of the boundary rarely contains only irrelevant environment objects, which rarely influence the manipulation feasibility inside the target region. Specifically, we use a two-channels 100x100 depth image. The first channel saves only the depth image of the target box, which is located in the image center. The second channel contains the depth image of other neighbor bodies relative to the target box. Besides the object-centered depth image, an extra feature array is also required to record the missed information of the environment, which includes the relative position of the target region to the robot base ($x_{t}$, $y_{t}$, $z_{t}$). Given the object-centered image and the feature array, the scope of the current action can be determined.


\begin{figure}[t]
	\centering
	\includegraphics[width=1\linewidth]{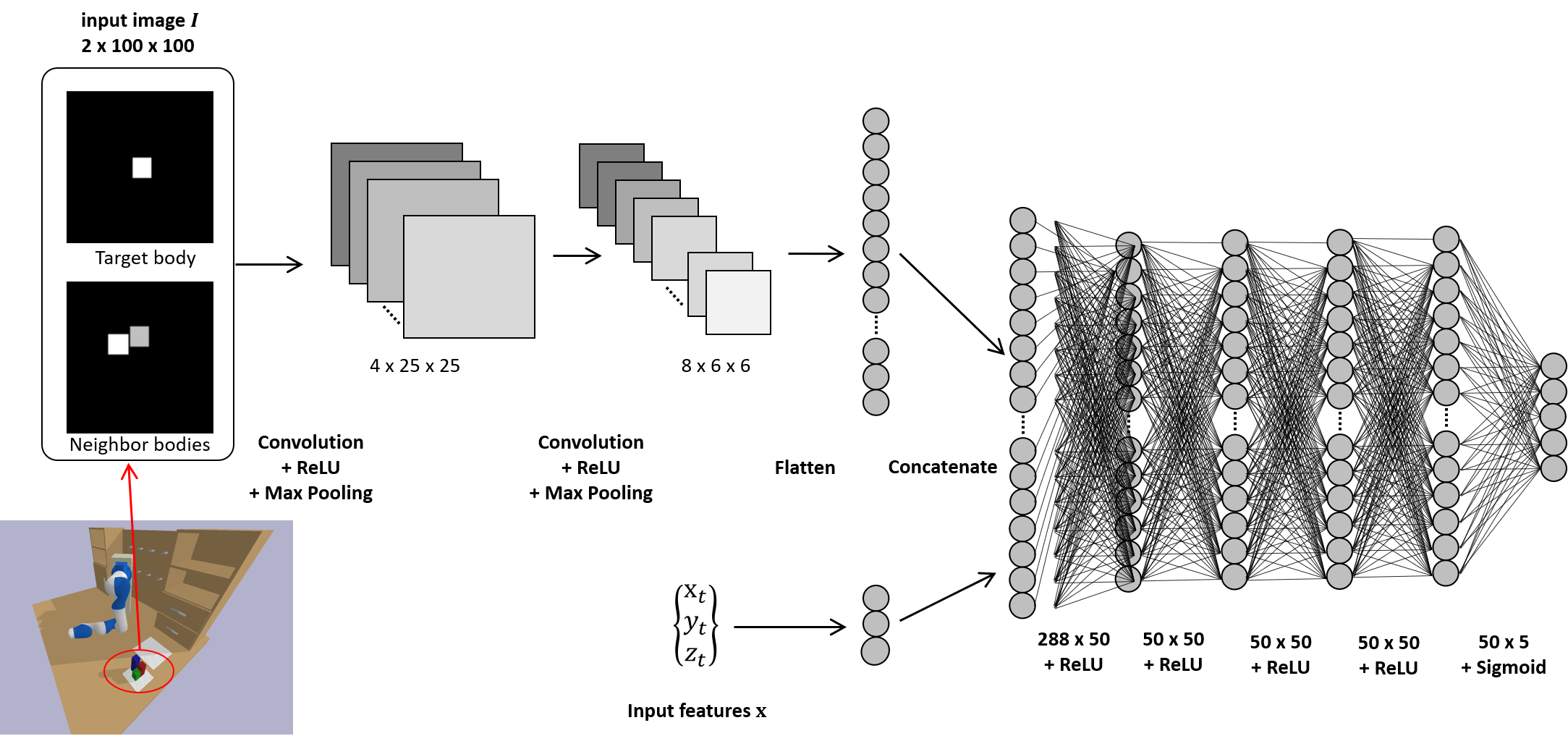}
	\caption{NFC model architecture. NFC model has two input data, a 2-channels depth images and a 3-D feature vector. Depth image contains the information of the target region, i.e. the size, shape and relative position of the target box and its neighbors. The feature vector contains the relative position of the region to the robot base. The output are 5-d feasibility probability in all 5 pick-up directions (see Fig. \ref{fig:visualiation pick direction}). We use two convolutional layers for extracting features, and extracted features and 3-d feature will be concatenated. Then 5 fully-connected dense layers will be used for changing the output dimension and learning the non-linear feasibility function.}
	\label{fig:cnn model structure}
\end{figure}

\subsection{Collecting Data}
\label{sec:nfc_data}
For the Pick-Place domain, the \textit{pick-up} action is to grasp one body with a specific grasping pose or \textit{put-down} is to place the box grasped in hand on the table in a specific location. \textit{put-down} action works just like the reversed action of \textit{pick-up} action. So we can view both types of action as the same one. Therefore, in the training process, we only need to collect the feasibility data about if the target body placed on table can be picked up from the initial robot configuration $q_0$. The feasibility of an action may be influenced in three aspects: inverse kinematics, collision between the robot and the other box, collision between the robot and the table. This means the learned classifier can help TAMP from consuming time on all three aspects.

In the training scene, it's impossible to consider the arbitrary number of objects in the workplace. So we train NFC with at least one body (the target body) and multiple box-shaped bodies (at most four bodies and three for neighboring bodies) inside the target region. One-body means only the target body inside the target region and the motion feasibility depends only on the collision between the robot and the table, i.e. the inverse kinematics, and the robot reachability. Multiple bodies means the neighbor body or neighbor bodies inside the region may also influence the manipulation feasibility of the target body. 


The position of the table and size of both bodies are uniformly randomly sampled. When sampling the position of neighbor bodies, we favor to sample neighbor bodies near to the target body, to generate more feasibility data of this critical case and to ensure this case won't be ignored during the model training. We represent every sampled scene with a object-centered image and feature array and record the feasibility labels of all 5 pick-up directions. 



\subsection{Training Model}
\label{sec:nfc_model}

The specific network architecture (see Fig. \ref{fig:cnn model structure}) used in the experiments: The CNN part is composed of two convolutional layers with filter in the size of 3 x 3, each followed by a max-pooling layer of size 2 and stride 2. The first convolution layer has 4 filter channels, the second one has 8. The output of the CNN is flattened to a feature in the size of 288. The extra feature size is 3. Both features are concatenated and passed through 4 additional fully connected hidden layers each with 50 neurons and output layers with 5 neurons. Except the final layer with one sigmoid output, all other hidden layers use ReLUs as the activation function.

We generate a dataset in the size of 240,000. 21.4\% of which are feasible cases and 78.6\% infeasible and more precisely 17.5\%,16.6\%,17.2\%,17.7\%, 38.2\% of which separately in 5 pick-up directions are feasible. We use Adam as the optimizer for model fitting (learning rate 0.001, batch size 32). To account for the data imbalance in the training dataset and different imbalance degree in different pick-up direction, we choose different loss weights (i.e. 4.7, 5.0, 4.8, 4.6 and 1.6) for different directions. 90\% of data is used for training the NFC model. When we choose 0.5 as the feasibility threshold (prediction probability >0.5 is feasible). The model prediction accuracy on test data is 93.6\%, with the false feasible rate of 3.65\% and the false infeasible rate of 2.74\%.

\subsection{Guiding Task and Motion Planning using NFC}
\label{sec:nfc_guiding_tamp}
We integrate our NFC into IDTMP to guide motion planning process. The workflow is shown in Fig. \ref{fig:IDTMP-NFC structure}. The only difference in comparison with the original IDTMP is that the NFC will predict the feasibility probability of each action in the task plan before the motion planner. The action with prediction probability greater than the feasibility threshold $\beta$, which is $>0.5$, will be considered as feasible. The failed step of infeasible plan will be directly returned. Therefore, NFC helps to rule out most infeasible action sequences for motion planner, so to save the most of motion planning time. Only NFC-feasible plan will be passed to time-consuming motion planner for re-checking and generating specific path.
   \section{EMPIRICAL EVALUATION}
\label{sec:evaluation}
We demonstrate our approach on the pick-place domain using KUKA collaborate robot arm. In the Section \ref{sec:idtmp_vs_nfc} we will show the remarkable improvement using NFC to guide the motion planning process of IDTMP. In the Section \ref{sec:NFC_vs_DVH} we compare our NFC with its predecessor, deep visual heuristics (DVH) \cite{driess_deep_2020} in aspects of training process. In the Section \ref{NFC_3D_space}, we will show how our approach generalizes and performs on 3-D Cartesian space.

\begin{figure}[t]
	\centering
	\includegraphics[width=1\linewidth]{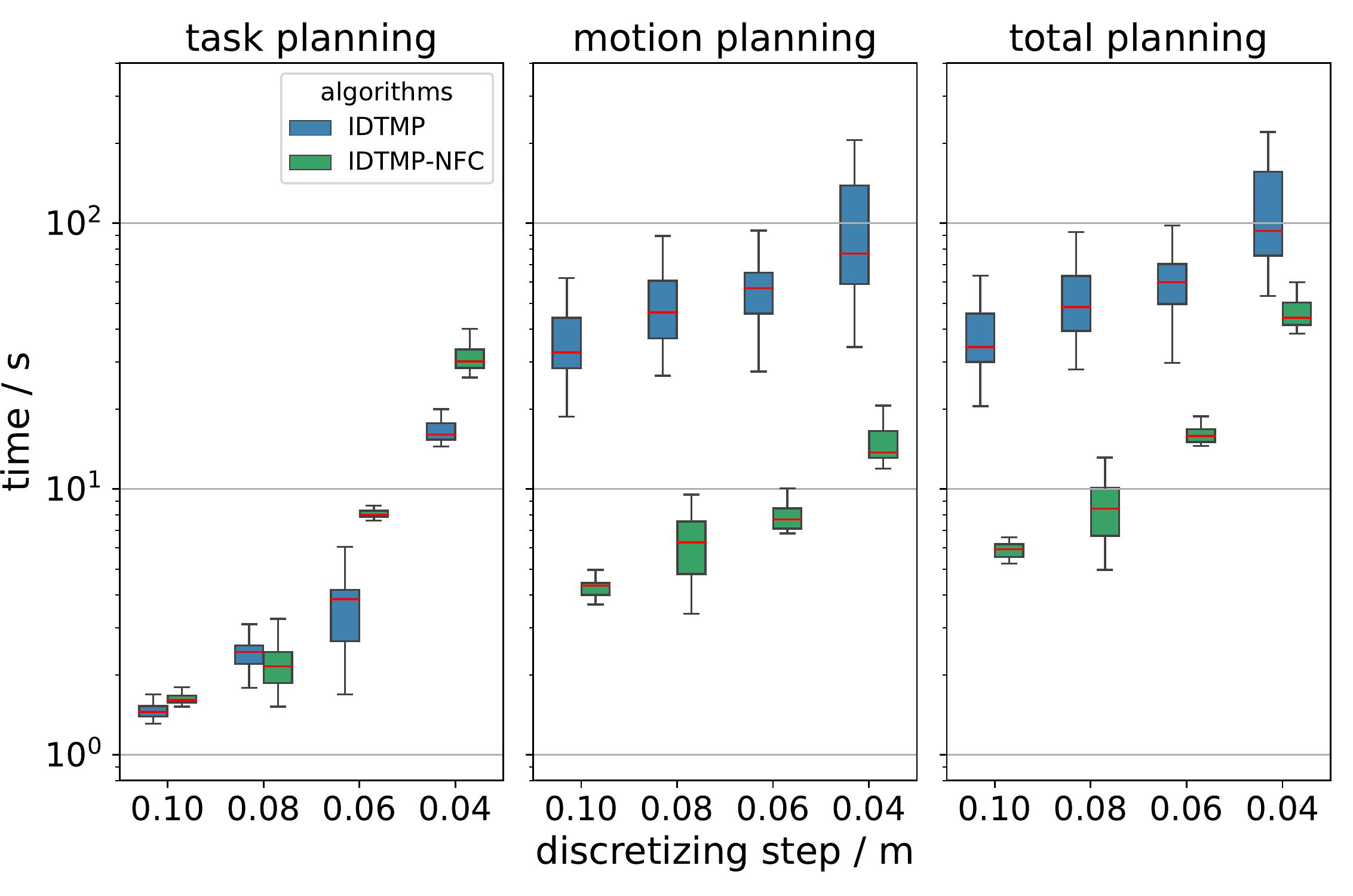}
	\caption{Comparison of the IDTMP and IDTMP using NFC on \textit{Unpack} problem in different discretizing steps. Motion planning time and total planning time are improved greatly. Task planning time increases slightly for discretizing step in 0.06 and 0.04 because false negative predictions lead to some correct task plans missed.}
	\label{fig:idtmp_vs_idtmp-nfc_unpack}
\end{figure}

\subsection{IDTMP-NFC vs. IDTMP}
\label{sec:idtmp_vs_nfc}
As mentioned in Section \ref{sec:pre_idtmp}, IDTMP needs first to pre-discretize the continuous geometric space into discrete space. We test and compare the original IDTMP and the IDTMP integrated with NFC in different discretizing step on \textit{Unpack} problem (see the problem description in Fig. \ref{fig:unpack scene}). A smaller discretizing step means larger geometric space and task space to search to try, e.g. there will be more candidate placing positions on the table for objects to be placed. From Fig. \ref{fig:idtmp_vs_idtmp-nfc_unpack}, we can conclude that NFC reduces the motion planning time of TAMP on the order of magnitudes and hence improves the TAMP process through guiding its motion planning tremendously.

\subsection{NFC vs. DVH}
\label{sec:NFC_vs_DVH}
DVH \cite{driess_deep_2020} uses the global depth image which captures the environment of the entire desktop. Our NFC uses object-centered depth image, which only focuses on the local environment around the target area in the size of 0.5m*0.5m. To ensure the same fineness of image, the resolution size of the global image of DVH will be much bigger than that of our local image and thus occupies much more RAM and requiring more time for the training process (see TABLE \ref{fig:nfc_vs_dvh}). Another remarkable benefit of NFC over DVH is, DVH is trained only for a specific scenario and is more like a single-use model. When the robot operates objects on a different desk or between several desks in different heights, another DVH model or even several separate DVH models are required.

\begin{table}[t]
\caption{Comparison Between Our NFC and DVH}
\label{fig:nfc_vs_dvh}
\begin{center}
\begin{tabular}{c|ccc}
\hline
                                                                          & \begin{tabular}[c]{@{}c@{}}NFC\\ 0.5*0.5\end{tabular} & \begin{tabular}[c]{@{}c@{}}DVH\\ 0.8*0.6\end{tabular} & \begin{tabular}[c]{@{}c@{}}DVH\\ 1.6*1.2\end{tabular} \\ \hline
\begin{tabular}[c]{@{}c@{}}image\\ resolution\end{tabular}                & 100*100                                               & 160*120                                               & 320*240                                               \\ \hline
\begin{tabular}[c]{@{}c@{}}RAM usage \\ per 10k data / GB\end{tabular}     & 0.19                                                  & 0.36                                                  & 1.43                                                  \\ \hline
\begin{tabular}[c]{@{}c@{}}Querying\\ time / ms\end{tabular}              & 0.038                                                 & 0.045                                                 & 0.092                                                 \\ \hline
\begin{tabular}[c]{@{}c@{}}number of model\\ parameters\end{tabular}      & 28,601                                                & 42,201                                                & 134,201                                               \\ \hline
\begin{tabular}[c]{@{}c@{}}training time \\ per 10k data / s\end{tabular} & 3.0                                                   & 5.2                                                   & 22.5                                                  \\ \hline
\end{tabular}
\end{center}
\end{table}



\subsection{NFC on the Entire Workspace}
\label{NFC_3D_space}
In order to test the effectiveness of our trained NFC model in the entire 3-d Cartesian space, we randomly sample 100 different scenes (random positions of two regions) for solving \textit{Unpack} problem, (see Fig. \ref{fig:random_unpack_scenes}). All the \textit{Unpack} problems are solvable with the original IDTMP. With NFC only 2 motion planning problems have to be solved rather than 26 on average for IDTMP without NFC. 63.5\% of the motion planning time is saved. Only 8\% of the scenes are infeasible for IDTMP-NFC due to the false infeasible prediction of the NFC model.

\begin{figure}[t]
\centering
	\includegraphics[width=1\linewidth]{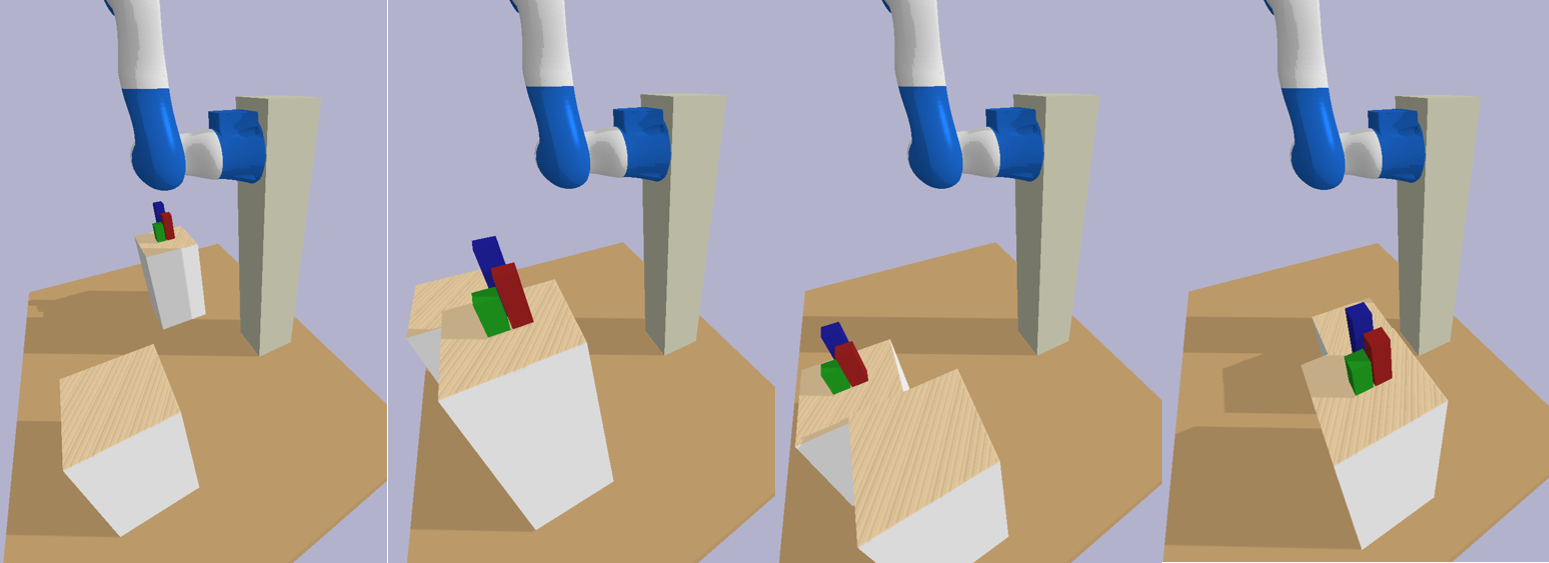}
	\caption{Different \textit{Unpack} problems. We randomly sample 100 different \textit{Unpack} problems with different table positions and test the effectiveness of our NFC model.}
	\label{fig:random_unpack_scenes}
\end{figure}
   \section{CONCLUSION}
\label{sec:conclude}
We propose a novel methods which enables to predict motion feasibility in the entire workspace. In the above demonstration, 
we have shown that NFC first improve the performance of the original TAMP on the order of magnitudes. Our NFC using object-centered images outperforms DVH in aspects of training process(e.g. training time and RAM usage) and model querying time. NFC shows good generalization and high prediction accuracy in the entire robot workspace. So once we have finished training the model, we can apply the model to all working scenarios of the current robot, not restricted to a specific desktop scenario.


For future work, we can add more features to increase the functionality of our model, e.g. to predict manipulation feasibility with/without obstacle overhead of the target body, so that the NFC model can deal with more general situations in real-world applications.
}

\begin{abstract}
As robots play an increasingly important role in the industrial, the expectations about their applications for everyday living tasks are getting higher. Robots need to perform long-horizon tasks that consist of several sub-tasks that need to be accomplished. \gls{tamp} provides a hierarchical framework to handle the sequential nature of manipulation tasks by interleaving a symbolic task planner that generates a possible action sequence, with a motion planner that checks the kinematic feasibility in the geometric world, generating robot trajectories if several constraints are satisfied, e.g., a collision-free trajectory from one state to another. Hence, the reasoning about the task plan's geometric grounding is taken over by the motion planner. However, motion planning is computationally intense and is usability as feasibility checker casts \gls{tamp} methods inapplicable to real-world scenarios. In this paper, we introduce \gls{nfc}, a simple yet effective visual heuristic for classifying the feasibility of proposed actions in \gls{tamp}. Namely, \gls{nfc} will identify infeasible actions of the task planner without the need for costly motion planning, hence reducing planning time in multi-step manipulation tasks. NFC encodes the image of the robot's workspace into a feature map thanks to \gls{cnn}. We train \gls{nfc} using simulated data from \gls{tamp} problems and label the instances based on IK feasibility checking. Our empirical results in different simulated manipulation tasks show that our \gls{nfc} generalizes to the entire robot workspace and has high prediction accuracy even in scenes with multiple obstructions. When combined with state-of-the-art integrated \gls{tamp}, our \gls{nfc} enhances its performance while reducing its planning time.

\end{abstract}

   \section{INTRODUCTION}
\label{sec:intro}

The application of autonomous robotics is rapidly increasing in the industrial and manufacturing sectors. However, a long-wished vision is having robots operate in real-world unstructured environments, interact with humans, and execute complex tasks that concern services in houses, nursing homes, hospitals, etc. While robots manifested repetitive tasks in static and structured settings, we are far from achieving the goal of general-purpose autonomous personal robots. One of the many challenges that hinder robots from executing complex tasks is their long planning time. Depending on the combinatorics of the problem that needs to be solved, planning may terminate without a solution \cite{garrett_integrated_2021}. 

While \gls{tamp} offers a framework for optimizing over task action-sequences and feasible geometric motions, it suffers from the logical gap between the symbolic and geometric space \cite{kaelbling_integrated_2013}. Indeed, in the early stages of the \gls{tamp} research, this hierarchical problem was decomposed into two separate sub-problems: first finding an action sequence and then finding continuous parameter values for each action. This strictly hierarchical approach is incomplete, as there are no guarantees that actions selected by the task planner will be geometrically feasible. For example, the Shakey robot executed STRIPS \cite{fikes_strips_1971} to plan a high-level abstract action plan, such as which room to move to, and then planned for the corresponding low-level motion. Unfortunately, if the low-level motion planning was not feasible, there was no mechanism to find an alternative high-level plan. This isolation of the planning problems in practice rarely holds. For example, in Fig. \ref{fig:unpack scene}, the robot has to move the green box to the other region. The straightforward action sequence is to pick up the green box and put it down to the other region. However, the pick-up action is blocked by the other two taller boxes thus, the low-level motion fails.


\begin{figure}[t]
	\centering
	\includegraphics[width=0.9\linewidth]{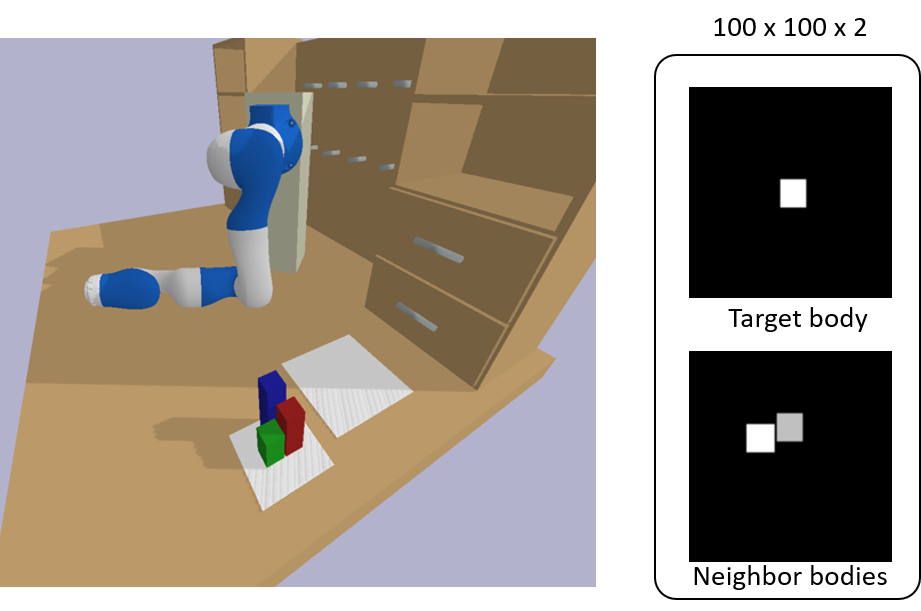}
	\caption{Scene of \textit{Unpack} problem. Left: Three boxes of different sizes are placed together. The goal is to place the green one in the other region. The robot is only allowed to pick a box from the above. TAMP solver must figure out that it has to relocate the two taller boxes away to reach the target green box. Right: We use the local depth image around the target body to predict the feasibility of an action. The depth image is composed of two channels, with the target box in the first channel and other boxes in the second channel.}
	\label{fig:unpack scene}
	\end{figure}

Over the last two decades, researchers naturally opted for extending \gls{tamp} considering the integration of discrete-continuous actions \cite{kaelbling_hierarchical_2011}, leading to a hybrid hierarchical search problem that involves, on the high-level task planning, i.e., the selection of a finite sequence or skeletons of discrete symbolic actions (e.g., which objects to pick and place), on the low-level motion planning, i.e., the selection of parameters in the continuous space to instantiate the sequence (e.g., the pose and grasp of the movable objects, path between two configurations), and in-between an interface for sharing information between discrete and continuous layer, as to trigger the generation of alternate skeletons when the motion generation of the current task plan fails \cite{garrett_integrated_2021,srivastava_combined_2014}. 

Interleaving task planning and motion planning, with the scope of finding a feasible solution, usually involves a large search space that is computationally exhaustive. In particular, the low-level motion planner has to validate the feasibility of the generated action sequences and to instantiate the final kinematically feasible action sequence, while the most action sequences are infeasible, which consumes much of the planning time, making most TAMP methods inefficient and impractical for real world scenario. To speed up the planning process in \gls{tamp} and increase its probability of finding an overall feasible solution in a given time-budget, several heuristics have been introduced to reduce the search space. In particular, in the era of robot learning, different learned heuristics have been integrated in \gls{tamp} to accelerate its decision-making process, and support its probabilistic completeness to find a feasible plan in a specific planning horizon \cite{wells_learning_2019, driess_deep_2020,silver2021learning}.  \cite{driess_deep_2020} introduces a more general visual-based neural network to do the classification of tabletop manipulation. In this paper, we propose a novel CNN-based neural feasibility classifier (NFC) using the local visual information around the target body, object-centered image. With this idea, we are able to extend the feasibility prediction to the entire workspace of robot manipulators. Our model is also able to deal with the case that there is obstacle over the target body.

We implement our work based on the sequencing-first and discretizing-based TAMP methods, iteratively deepened task and motion planning (IDTMP). The main contributions in this work are two-fold,
\begin{itemize}
\item We propose a more general neural feasibility classifier (NFC) that predicts the feasibility of action based on object-centered image of the scene.
\item We show the generalization and high prediction accuracy of the NFC in the entire workspace of the manipulator.
\end{itemize}
   \section{RELTAED WORK}
\label{sec:related_work}

\subsection{TAMP}
\label{sec:rw_tamp}

Theoretically, general-purpose TAMP uses symbolic AI planning \cite{ren_extended_2021} to search in task spaces and can deal with problems with large action space, non-geometric action, etc. Top-down hierarchical planning means also sequencing-first planning \cite{garrett_integrated_2021}. It generates high-level task plan first and maps each action of the plan to the motion space and finds the corresponding motion plan. Sequencing-first planning is more in line with human behavior. In this paper, we only talk about the related works in this direction. Sequencing-first TAMP methods are time efficient while the time-consuming motion planning process is postponed as late as possible \cite{garrett_integrated_2021}.

Due to the inherent difficulty of task planning and motion planning, the former is NP-hard, and the latter is PSPACE hard. There is a compromiss between algorithm performance and algorithm completeness. TAMP methods in the early stage consider only restricted domains to improve the efficient \cite{strip_shakey_1966}, \cite{stilman_navigation_nodate}, \cite{cambon_robot_nodate}. \cite{kaelbling_hierarchical_2011} can guarantee completeness under the assumption of reversible action but at the cost of large amount of time and energy consuming. Therefore, the second challenge of algorithm design for integrated TAMP besides the completeness is to combine task planner and motion planner efficiently. There are two typical strategies to deal with the hybrid planning of the discrete task space and continuous motion space, discretizing-based and sampling-based. \cite{erdem_combining_2011}, \cite{lagriffoul_efficiently_2014}, \cite{lozano-perez_constraint-based_2014}, \cite{dantam_incremental_2018} require the pre-discretization of the continuous space so that the solution space can be interactively pruned according to the motion constraints from motion planning. \cite{srivastava_combined_2014}, \cite{ren_extended_2021}, \cite{garrett_pddlstream_2020} directly sample in the continuous space during the planning process, to guarantee that no potential poses will be missed by the discretization process and show better robustness. However, both strategies need to continuously call low-level motion planner to exclude most infeasible high-level task plans. Motion planner becomes the bottleneck of the original TAMP methods, since it have to be called so many times, to validate the feasibility of candidate task plans.

\subsection{Heuristic for Motion Planning}
\label{sec:rw_heuristics_mp}
In order to improve the performance of motion planning, there are many tricks and techniques being proposed. E-graph \cite{phillips_e-graphs_2012} which is learned from the online planning experience, represents the connectivity of the space and accelerates its planning effort, but it's not clear how it should be adapted to scenes with unknown objects. Lightning \cite{berenson_robot_2012} and Thunder \cite{coleman_experience-based_2014} are two experience-based frameworks using the strategy of experience storing, retrieving, and repairing to speed up high-dimensional motion planning problems. \cite{arslan_machine_2015} proposes a learning-based heuristic to predict the collision-free sample and to estimate the relevant region of the problem. \cite{ichter_learning_2019} proposes a methodology for non-uniform sampling using the sampling distribution learned from demonstration, i.e. successful motion planning, human demo. Our approach differs by using learning to guide the TAMP search and doesn't require any modification of the motion planner. Therefore, above mentioned heuristics for motion planner can also be viewed as a complementary to our method.

\subsection{Learning to Guide TAMP}
\label{sec:rw_heuristics_tamp}
There are also learning-based techniques, which aim to guide the TAMP process. \cite{chitnis_guided_2016} trains heuristics to estimate the difficulty of high-level plan refinement and uses reinforcement learning to propose continuous value for plan refinement in low-level, but human demonstration is required for training and detail about prediction error is not mentioned. “Score-space” is introduced in \cite{kim_learning_2019} as a metric to measure similarity between problem instances to improve motion planning time and transfers knowledge to other problems. Dex-net \cite{mahler_dex-net_2017} can predict the success rate for picking up an object with an arbitrary shape. But the interference of neighbor bodies is not considered. Closely related to our work in this paper, Wells et. al. \cite{wells_learning_2019} trained a feature-based SVM model to help with checking the feasibility of actions instead of using the expensive motion planner. For the same purpose, \cite{driess_deep_2020} validate that it's possible to predict the motion feasibility of a mixed-integer program from visual input with a high accuracy. The learned DVH model can well generalize to scenes with multiple objects and object with different shapes rather than Wells' feature-based SVM model. However, Both ideas are only applicable for a manipulator to operate on a desktop scenario. Our approach will overcome this problem and predict the feasibility of actions in the entire 3D-Cartesian workspace of a specific robot.

   \section{PRELIMINARIES}
\label{sec:preliminary}

\subsection{Task Planning}
\label{sec:pre_tp}
Task planning finds a discrete sequence of actions to transition from the given start state to the desired goal condition. For the general-purpose TAMP problem, PDDL (Planning Domain Definition Language) is the most widely used modeling language to describe the planning domain and a instantiated planning problem. PDDL-domain defines the “universal” aspects of problems. Essentially, these are the aspects that do not change regardless of what specific situation we’re trying to solve. A typical PDDL-domain file contains,
\begin{itemize}
    \item \textbf{predicates} $S$: the symbolic states. The scene state can be described using the combination of predicates.
    \item \textbf{operators} $A$: the symbolic actions, describing the transition of two states, which mainly includes action preconditions and effects after taking the action.
\end{itemize}

We also need a PDDL-problem file to model a specific planning problem. The PDDL-problem describes the following elements:
\begin{itemize}
    \item \textbf{initial states} $s_{0}$ described with the combination of predicates.
    \item \textbf{a set of goal states} $S_g$ described with the combination of predicates.
\end{itemize}

With the PDDL-domain and PDDL-problem defined, there are many task planner $\mathcal{TP}$ in AI planning community available for generating task plans. A task plan $T$ or action sequence $<a_0, ..., a_{n-1}>$ transits the initial state $s_0$ sequentially to one state in the goal state set $S_g$. So the task planning problem can be formulated as (1).
$$
T = <a_0, ..., a_{n-1}> = \mathcal{TP}(S,A,s_0,S_g) \eqno{(1)}
$$
such that for, $ \forall i=0, 1, ... n-1 ,\\ $
$$
    s_i + a_i = s_{i+1} ,
    a_i \in A ,
    s_i \in S ,
    s_n \in S_g 
$$

\subsection{Motion Planning}
\label{sec:pre_mp}
Motion planning for manipulator $\mathcal{MP}$ is to find a motion plan $M$ or configuration sequence $<q_0, ..., q_n>$ which transfers the robot from its initial configuration $q_0$ to the one of the goal configuration $q_n$ without self-collision and collision with the environment, formulated in (2). No collision also means that every configuration of the motion planning solution should be located in the free configuration space of the manipulator $C_{free}$.

$$
    M = <q_0, ..., q_n> = \mathcal{MP}(q_0, q_n, C_{free}) \eqno{(2)}
$$

\subsection{Task and Motion Planning}
\label{sec:pre_tamp}
The key requirement for task planner in TAMP is to generate alternate high-level plans. As TAMP iterates between the task planning and motion planning layers, feedback from the motion planner ideally influences the task planner to generate new plans or to disapprove of geometrically difficult or even infeasible plans. Therefore, the task planning layer must be able to calculate alternative plans and ideally reuse previous planning experience to improve performance.

TAMP combines the discrete action selection of task planning with the continuous path generation of motion planning. The fundamental requirement of the task–motion layer is to build a bridge between task operators and motion planning problems. Given a symbolic task plan, what are the corresponding geometric plan? How should task planner react to the feedback from motion planner? The task–motion interface must translate between the low-level scene geometry and the high-level task descriptions.

We can formulate general TAMP problems as (3)
$$
T, M = \mathcal{TAMP}(S, A, s_0, S_g, C_{free}, q_{0}) \eqno{(3)}
$$

TAMP generates a sequence of actions $T = (a_0, a_1, ..., a_{n-1})$ following a sequence of task states $s_0, s_1, ..., s_n$ and a sequence to motion plans $M_0, M_1, ..., M_{n-1}$, such that for $\forall i= 0, 1, ..., n-1$, \\
$$
    s_n \in S_g,\\
    s_i + a_i = s_{i+1},\\
    a_i \Leftrightarrow M_i,\\
$$
$$
    M_i(0) \in s_i,\\
    M_i(1) \in s_{i+1},\\
    M_i(1) = M_{i+1}(0)\\
$$

\begin{figure}[t]
	\centering
	\includegraphics[width=1\linewidth]{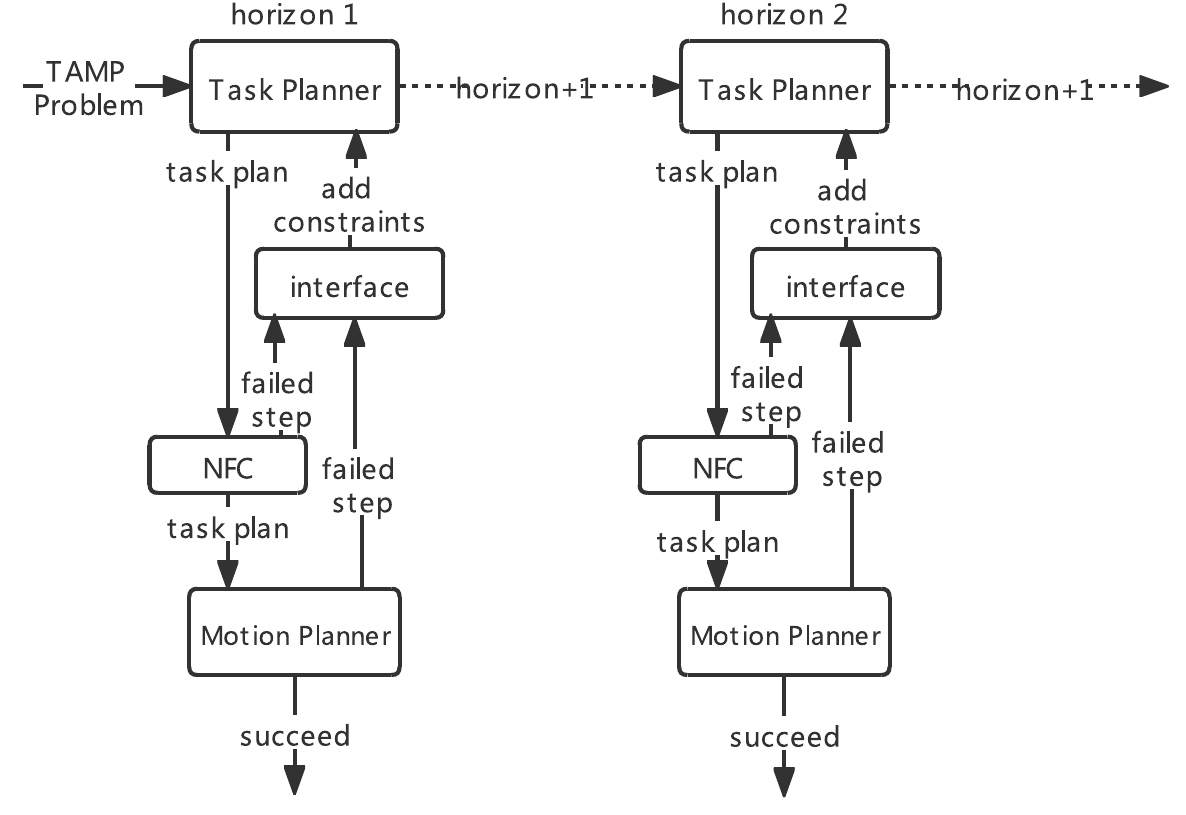}
	\caption{The Workflow of IDTMP integrated with NFC. The original IDTMP doesn't have NFC and motion planner must classify the feasibility of each action sequence by itself. As for IDTMP-NFC, only feasible task plan validated by NFC will be passed to the time-consuming motion planner. As for infeasible task plans, NFC will return its failed step to the interface for updating motion constraints.}
	\label{fig:IDTMP-NFC structure}
\end{figure}

\subsection{Iteratively Deepened Task and Motion Planning (IDTMP)}
\label{sec:pre_idtmp}
Theoretically, our NFC is applicable to all versions of sequencing-first TAMP methods. We take the discretizing-based IDTMP \cite{dantam_incremental_2018} \cite{dantam_task-motion_2018} as the example and integrate our NFC with it. The algorithm workflow of IDTMP is shown in Fig. \ref{fig:IDTMP-NFC structure}. As for its task planner, it employs constraint-based SAT-planner to generate single candidate task plan according to the logical and motion constraints iteratively. Motion planner, such as sampling-based rapidly-exploring random tree(RRT) will validate the feasibility of each action in the candidate task plan. When it's infeasible, motion planner will return failed step so that the interface layer can add extra motion constraints to SAT-planner, so to prune the symbolic solution space. When feasible, motion planner will find the corresponding geometric motion for each action, such as inverse kinematics and path in configuration space. IDTMP searches and validates plans incrementally or horizon by horizon. When task and motion planning on the short horizon failed, IDTMP will jump onto longer horizons and try more complex task plans. To ensure the algorithm completeness, IDTMP will clear all collected motion constraints on the last horizon, so that previously failed task plans could be retrospected. TAMP process come to end until both task and motion plan are found or the allotted planning time is exceeded.


   \section{NEURAL FEASIBILITY CLASSIFIER}
\label{sec:nfc}
The goal of our work is to train a classifier which can predict the action feasibility in the 3d-Cartesian workspace of a specific robot arm. In order to realize this goal, the key requirement is how to encode involved elements, including the target object of different shape, the environment and action type. Driess et. al. \cite{driess_deep_2020}, \cite{driess_reasoning_2020} have shown the effectiveness of visual heuristics on desktop scenarios. Our heuristic is also visual-based but applies a more efficient way to encode the scene.

\begin{figure}[t]
    \centering
    \includegraphics[width=1\linewidth]{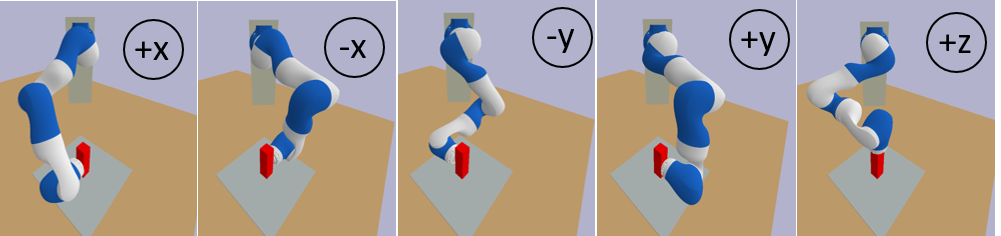}
    \caption{Visualization of 5 different discrete actions. Different actions means it has different pick-up or put-down directions, which are +x, -x, +y, -y, +z direction of the body frame.}
    \label{fig:visualiation pick direction}
\end{figure}

Our neural feasibility classifier employs a local depth image (Fig. \ref{fig:unpack scene}) which captures the neighboring area (boundary of 0.5m x 0.5m) around the target body, object-centered image. The neighboring area of target body means, the area around to be picked object for \textit{pick-up} action, and the area around the placing pose of grasped body for \textit{put-down} action. The image pixel out of the boundary rarely contains only irrelevant environment objects, which rarely influence the manipulation feasibility inside the target region. Specifically, we use a two-channels 100x100 depth image. The first channel saves only the depth image of the target box, which is located in the image center. The second channel contains the depth image of other neighbor bodies relative to the target box. Besides the object-centered depth image, an extra feature array is also required to record the missed information of the environment, which includes the relative position of the target region to the robot base ($x_{t}$, $y_{t}$, $z_{t}$). Given the object-centered image and the feature array, the scope of the current action can be determined.


\begin{figure}[t]
	\centering
	\includegraphics[width=1\linewidth]{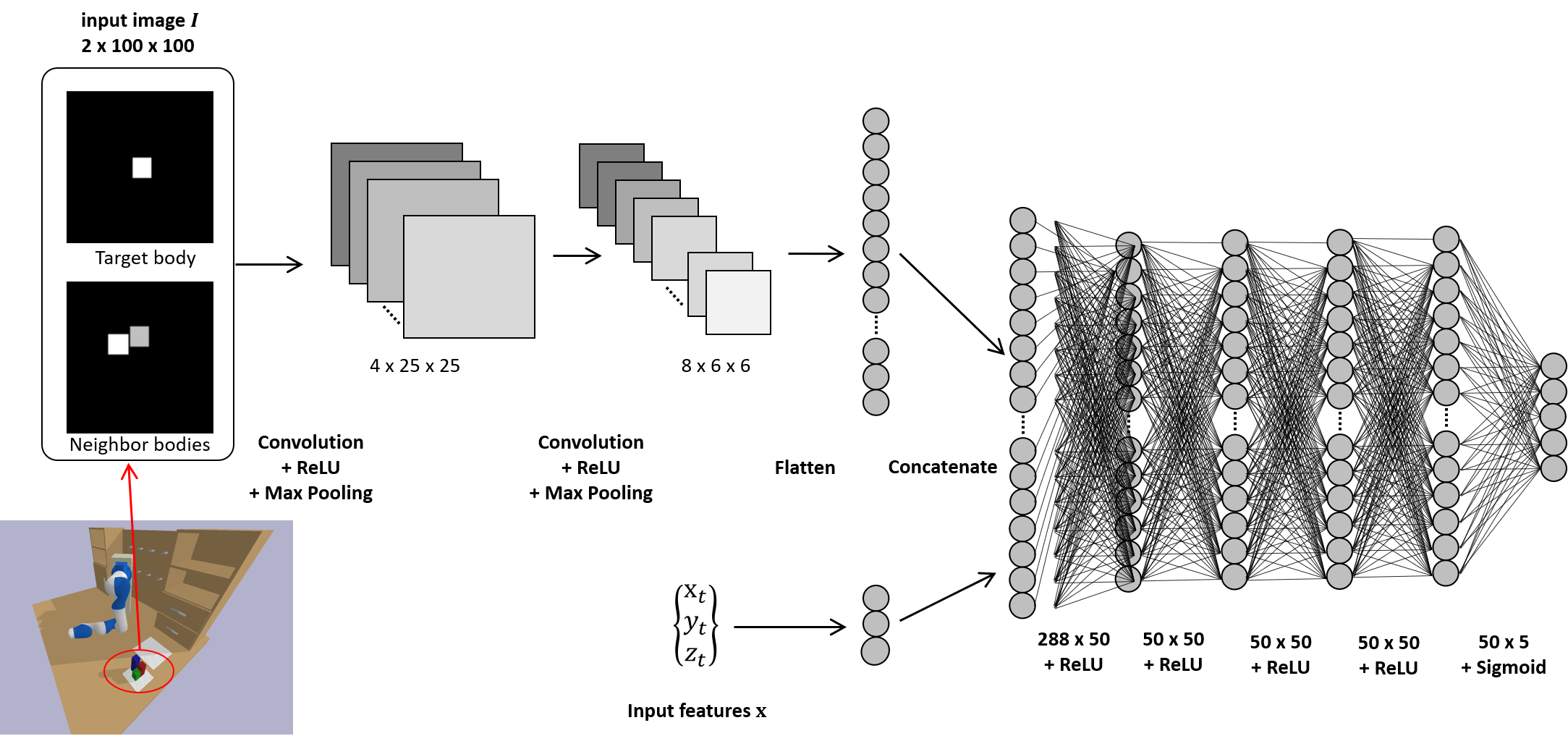}
	\caption{NFC model architecture. NFC model has two input data, a 2-channels depth images and a 3-D feature vector. Depth image contains the information of the target region, i.e. the size, shape and relative position of the target box and its neighbors. The feature vector contains the relative position of the region to the robot base. The output are 5-d feasibility probability in all 5 pick-up directions (see Fig. \ref{fig:visualiation pick direction}). We use two convolutional layers for extracting features, and extracted features and 3-d feature will be concatenated. Then 5 fully-connected dense layers will be used for changing the output dimension and learning the non-linear feasibility function.}
	\label{fig:cnn model structure}
\end{figure}

\subsection{Collecting Data}
\label{sec:nfc_data}
For the Pick-Place domain, the \textit{pick-up} action is to grasp one body with a specific grasping pose or \textit{put-down} is to place the box grasped in hand on the table in a specific location. \textit{put-down} action works just like the reversed action of \textit{pick-up} action. So we can view both types of action as the same one. Therefore, in the training process, we only need to collect the feasibility data about if the target body placed on table can be picked up from the initial robot configuration $q_0$. The feasibility of an action may be influenced in three aspects: inverse kinematics, collision between the robot and the other box, collision between the robot and the table. This means the learned classifier can help TAMP from consuming time on all three aspects.

In the training scene, it's impossible to consider the arbitrary number of objects in the workplace. So we train NFC with at least one body (the target body) and multiple box-shaped bodies (at most four bodies and three for neighboring bodies) inside the target region. One-body means only the target body inside the target region and the motion feasibility depends only on the collision between the robot and the table, i.e. the inverse kinematics, and the robot reachability. Multiple bodies means the neighbor body or neighbor bodies inside the region may also influence the manipulation feasibility of the target body. 


The position of the table and size of both bodies are uniformly randomly sampled. When sampling the position of neighbor bodies, we favor to sample neighbor bodies near to the target body, to generate more feasibility data of this critical case and to ensure this case won't be ignored during the model training. We represent every sampled scene with a object-centered image and feature array and record the feasibility labels of all 5 pick-up directions. 



\subsection{Training Model}
\label{sec:nfc_model}

The specific network architecture (see Fig. \ref{fig:cnn model structure}) used in the experiments: The CNN part is composed of two convolutional layers with filter in the size of 3 x 3, each followed by a max-pooling layer of size 2 and stride 2. The first convolution layer has 4 filter channels, the second one has 8. The output of the CNN is flattened to a feature in the size of 288. The extra feature size is 3. Both features are concatenated and passed through 4 additional fully connected hidden layers each with 50 neurons and output layers with 5 neurons. Except the final layer with one sigmoid output, all other hidden layers use ReLUs as the activation function.

We generate a dataset in the size of 240,000. 21.4\% of which are feasible cases and 78.6\% infeasible and more precisely 17.5\%,16.6\%,17.2\%,17.7\%, 38.2\% of which separately in 5 pick-up directions are feasible. We use Adam as the optimizer for model fitting (learning rate 0.001, batch size 32). To account for the data imbalance in the training dataset and different imbalance degree in different pick-up direction, we choose different loss weights (i.e. 4.7, 5.0, 4.8, 4.6 and 1.6) for different directions. 90\% of data is used for training the NFC model. When we choose 0.5 as the feasibility threshold (prediction probability >0.5 is feasible). The model prediction accuracy on test data is 93.6\%, with the false feasible rate of 3.65\% and the false infeasible rate of 2.74\%.

\subsection{Guiding Task and Motion Planning using NFC}
\label{sec:nfc_guiding_tamp}
We integrate our NFC into IDTMP to guide motion planning process. The workflow is shown in Fig. \ref{fig:IDTMP-NFC structure}. The only difference in comparison with the original IDTMP is that the NFC will predict the feasibility probability of each action in the task plan before the motion planner. The action with prediction probability greater than the feasibility threshold $\beta$, which is $>0.5$, will be considered as feasible. The failed step of infeasible plan will be directly returned. Therefore, NFC helps to rule out most infeasible action sequences for motion planner, so to save the most of motion planning time. Only NFC-feasible plan will be passed to time-consuming motion planner for re-checking and generating specific path.
   \section{EMPIRICAL EVALUATION}
\label{sec:evaluation}
We demonstrate our approach on the pick-place domain using KUKA collaborate robot arm. In the Section \ref{sec:idtmp_vs_nfc} we will show the remarkable improvement using NFC to guide the motion planning process of IDTMP. In the Section \ref{sec:NFC_vs_DVH} we compare our NFC with its predecessor, deep visual heuristics (DVH) \cite{driess_deep_2020} in aspects of training process. In the Section \ref{NFC_3D_space}, we will show how our approach generalizes and performs on 3-D Cartesian space.

\begin{figure}[t]
	\centering
	\includegraphics[width=1\linewidth]{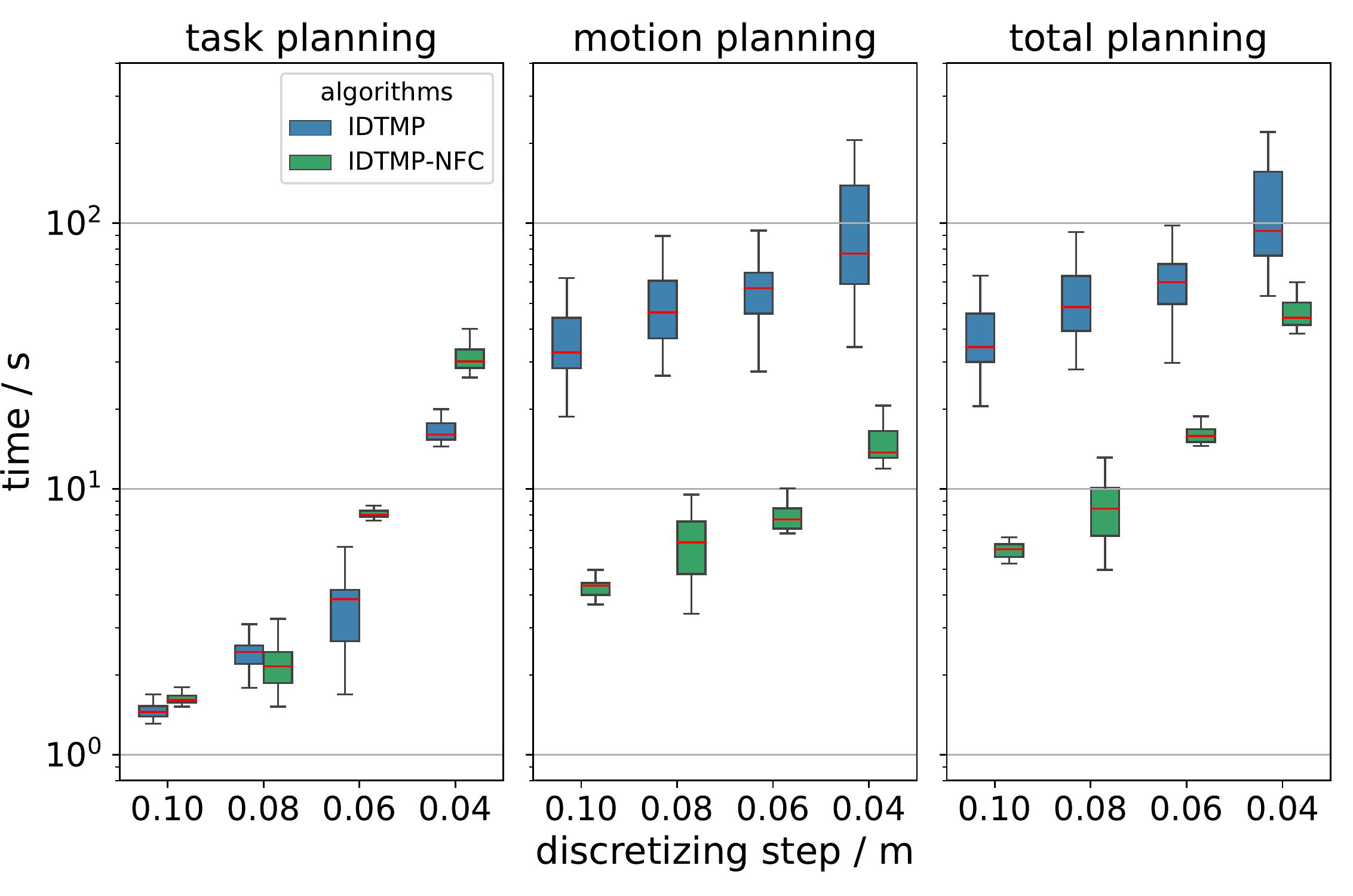}
	\caption{Comparison of the IDTMP and IDTMP using NFC on \textit{Unpack} problem in different discretizing steps. Motion planning time and total planning time are improved greatly. Task planning time increases slightly for discretizing step in 0.06 and 0.04 because false negative predictions lead to some correct task plans missed.}
	\label{fig:idtmp_vs_idtmp-nfc_unpack}
\end{figure}

\subsection{IDTMP-NFC vs. IDTMP}
\label{sec:idtmp_vs_nfc}
As mentioned in Section \ref{sec:pre_idtmp}, IDTMP needs first to pre-discretize the continuous geometric space into discrete space. We test and compare the original IDTMP and the IDTMP integrated with NFC in different discretizing step on \textit{Unpack} problem (see the problem description in Fig. \ref{fig:unpack scene}). A smaller discretizing step means larger geometric space and task space to search to try, e.g. there will be more candidate placing positions on the table for objects to be placed. From Fig. \ref{fig:idtmp_vs_idtmp-nfc_unpack}, we can conclude that NFC reduces the motion planning time of TAMP on the order of magnitudes and hence improves the TAMP process through guiding its motion planning tremendously.

\subsection{NFC vs. DVH}
\label{sec:NFC_vs_DVH}
DVH \cite{driess_deep_2020} uses the global depth image which captures the environment of the entire desktop. Our NFC uses object-centered depth image, which only focuses on the local environment around the target area in the size of 0.5m*0.5m. To ensure the same fineness of image, the resolution size of the global image of DVH will be much bigger than that of our local image and thus occupies much more RAM and requiring more time for the training process (see TABLE \ref{fig:nfc_vs_dvh}). Another remarkable benefit of NFC over DVH is, DVH is trained only for a specific scenario and is more like a single-use model. When the robot operates objects on a different desk or between several desks in different heights, another DVH model or even several separate DVH models are required.

\begin{table}[t]
\caption{Comparison Between Our NFC and DVH}
\label{fig:nfc_vs_dvh}
\begin{center}
\begin{tabular}{c|ccc}
\hline
                                                                          & \begin{tabular}[c]{@{}c@{}}NFC\\ 0.5*0.5\end{tabular} & \begin{tabular}[c]{@{}c@{}}DVH\\ 0.8*0.6\end{tabular} & \begin{tabular}[c]{@{}c@{}}DVH\\ 1.6*1.2\end{tabular} \\ \hline
\begin{tabular}[c]{@{}c@{}}image\\ resolution\end{tabular}                & 100*100                                               & 160*120                                               & 320*240                                               \\ \hline
\begin{tabular}[c]{@{}c@{}}RAM usage \\ per 10k data / GB\end{tabular}     & 0.19                                                  & 0.36                                                  & 1.43                                                  \\ \hline
\begin{tabular}[c]{@{}c@{}}Querying\\ time / ms\end{tabular}              & 0.038                                                 & 0.045                                                 & 0.092                                                 \\ \hline
\begin{tabular}[c]{@{}c@{}}number of model\\ parameters\end{tabular}      & 28,601                                                & 42,201                                                & 134,201                                               \\ \hline
\begin{tabular}[c]{@{}c@{}}training time \\ per 10k data / s\end{tabular} & 3.0                                                   & 5.2                                                   & 22.5                                                  \\ \hline
\end{tabular}
\end{center}
\end{table}



\subsection{NFC on the Entire Workspace}
\label{NFC_3D_space}
In order to test the effectiveness of our trained NFC model in the entire 3-d Cartesian space, we randomly sample 100 different scenes (random positions of two regions) for solving \textit{Unpack} problem, (see Fig. \ref{fig:random_unpack_scenes}). All the \textit{Unpack} problems are solvable with the original IDTMP. With NFC only 2 motion planning problems have to be solved rather than 26 on average for IDTMP without NFC. 63.5\% of the motion planning time is saved. Only 8\% of the scenes are infeasible for IDTMP-NFC due to the false infeasible prediction of the NFC model.

\begin{figure}[t]
\centering
	\includegraphics[width=1\linewidth]{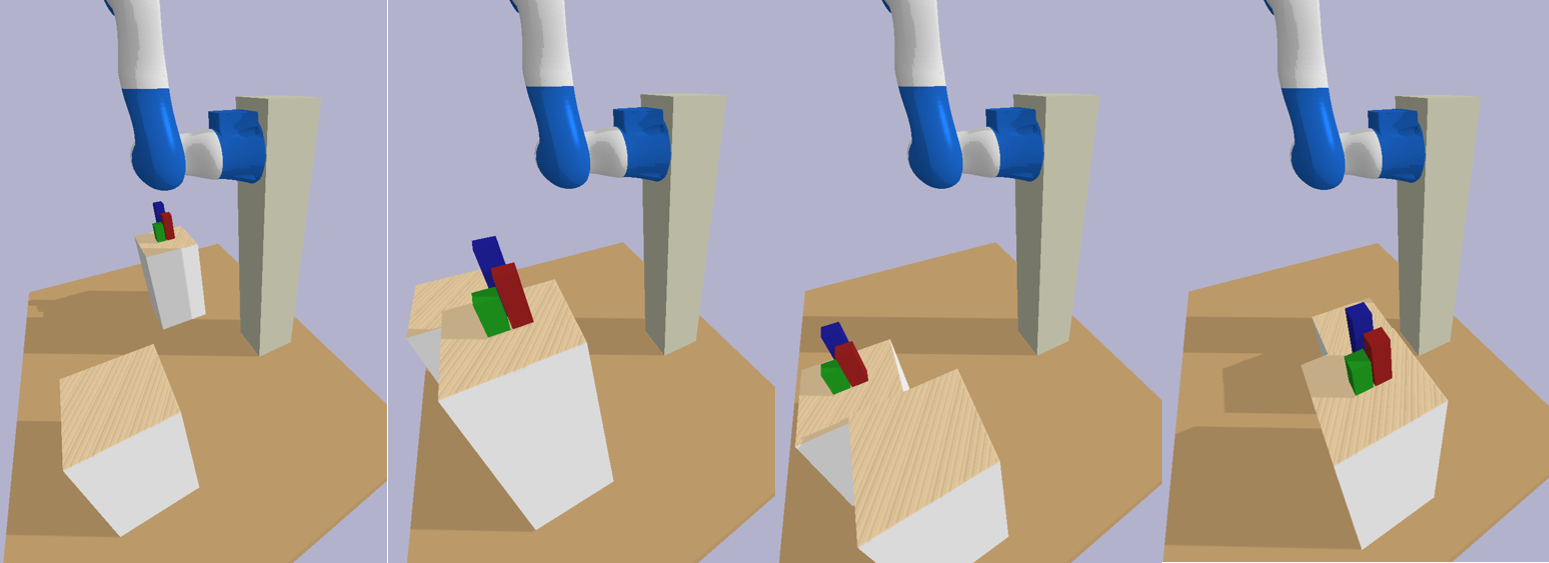}
	\caption{Different \textit{Unpack} problems. We randomly sample 100 different \textit{Unpack} problems with different table positions and test the effectiveness of our NFC model.}
	\label{fig:random_unpack_scenes}
\end{figure}
   \section{CONCLUSION}
\label{sec:conclude}
We propose a novel methods which enables to predict motion feasibility in the entire workspace. In the above demonstration, 
we have shown that NFC first improve the performance of the original TAMP on the order of magnitudes. Our NFC using object-centered images outperforms DVH in aspects of training process(e.g. training time and RAM usage) and model querying time. NFC shows good generalization and high prediction accuracy in the entire robot workspace. So once we have finished training the model, we can apply the model to all working scenarios of the current robot, not restricted to a specific desktop scenario.


For future work, we can add more features to increase the functionality of our model, e.g. to predict manipulation feasibility with/without obstacle overhead of the target body, so that the NFC model can deal with more general situations in real-world applications.

\bibliographystyle{IEEEtran}

\bibliography{IEEEtranBST/IEEEexample}

\end{document}